\documentclass[letterpaper, preprint, paper,11pt]{AAS}	
\usepackage[utf8]{inputenc}

\usepackage{graphicx}
\usepackage{amsmath}
\usepackage{amsfonts}
\usepackage[version=4]{mhchem}
\usepackage{siunitx}
\usepackage{longtable,tabularx}
\usepackage{booktabs}
\usepackage{adjustbox}
\usepackage{float}
\usepackage[tight,footnotesize]{subfigure}
\usepackage{appendix}
\usepackage{url}
\usepackage[section]{placeins}

\usepackage{nomencl}
\makenomenclature

\usepackage{xcolor}

\newcommand{\figref}[1]{Figure~\ref{#1}}
\newcommand{\tabref}[1]{Table~\ref{#1}}

\renewcommand{\eqref}[1]{Eq.~\ref{#1}}

\newcommand{\obs}[0]{\boldsymbol{o}}

\newcommand{\state}[0]{\boldsymbol{s}}

\newcommand{\action}[0]{\boldsymbol{u}}

\newcommand{\reward}[0]{r}

\newcommand{\policy}[0]{\pi}

\newcommand{\radius}[0]{d}
\newcommand{\azimuth}[0]{\theta_{a}}
\newcommand{\elevationangle}[0]{\theta_{e}}
\newcommand{\sunangle}[0]{\theta_{\rm Sun}}

\newcommand{\numpoints}[0]{n_p}

\newcommand{\totalvel}[0]{\Vert \boldsymbol{v} \Vert}

\newcommand{\position}[0]{\boldsymbol{p}}
\newcommand{\deltav}[0]{\Delta V}
\newcommand{\deltat}[0]{\Delta t}

\newcommand{\meanmotion}[0]{\eta}
\newcommand{\mass}[0]{m}
\newcommand{\ups}[0]{\overrightarrow{r}}

\title{
Investigating the Impact of Observation Space Design Choices On Training Reinforcement Learning Solutions for Spacecraft Problems 
}

\author{
Nathaniel Hamilton\thanks{AI Scientist, Intelligent Systems Division, Parallax Advanced Research, 4035 Colonel Glenn Hwy, Beavercreek, OH, 45431.},
Kyle Dunlap\thanks{Aerospace Engineer, Autonomy Capability Team (ACT3), Air Force Research Laboratory, 2241 Avionics Circle, Wright-Patterson AFB, OH, 45433.},
and Kerianne L. Hobbs\thanks{Safe Autonomy Lead, Autonomy Capability Team (ACT3), Air Force Research Laboratory, 2241 Avionics Circle, Wright-Patterson AFB, OH, 45433.}
}

\PaperNumber{25-147}

\begin{document}

\maketitle

\begin{abstract}
Recent research using Reinforcement Learning (RL) to learn autonomous control for spacecraft operations has shown great success. However, a recent study showed their performance could be improved by changing the action space, i.e. control outputs, used in the learning environment. This has opened the door for finding more improvements through further changes to the environment. The work in this paper focuses on how changes to the environment's observation space can impact the training and performance of RL agents learning the spacecraft inspection task. The studies are split into two groups. The first looks at the impact of sensors that were designed to help agents learn the task. The second looks at the impact of reference frames, reorienting the agent to see the world from a different perspective. The results show the sensors are not necessary, but most of them help agents learn more optimal behavior, and that the reference frame does not have a large impact, but is best kept consistent. 
\end{abstract}



\section{Introduction}

Autonomous spacecraft operation is a critical capability for managing the growing number of space and increasingly complex operations. For \textit{On-orbit Servicing, Assembly, and Manufacturing} (OSAM) missions, the inspection task enables the ability to assess, plan for, and execute different objectives. While this task is traditionally executed using classical control methods, that requires constant monitoring and adjustment by human operators, which becomes challenging or even impossible as the complexity of the task increases. Therefore, there is a growing need for developing high-performing autonomy.

\textit{Reinforcement Learning} (RL) is a fast-growing field for developing high-performing autonomy. RL's growth is spurred by success in agents that learn to beat human experts in games like Go \cite{silver2016mastering}, Starcraft \cite{starcraft2019}, and Gran Turismo \cite{wurman2022outracing}. RL is a promising application for spacecraft operations because the learned solutions are able to react in real-time to changing mission objectives and environmental uncertainty \cite{ravaioli2022safe, hamilton2022zero}. Previous work demonstrates RL learning policies capable of completing inspection missions using waypoints \cite{LeiGNC22, AurandGNC23} or direct thrust control \cite{vanWijkAAS_23, dunlap2025demonstrating}, and inspecting an uncooperative space object \cite{Brandonisio2021}. Additionally, recent work has demonstrated RL-trained controllers successfully transferring from simple simulation environments to running on real-world systems \cite{dunlap2025demonstrating} and those same controllers are capable of running at operational speeds on space-grade hardware \cite{hamilton2025space}.

Despite these successes, recent work in \cite{hamilton2024investigating} has demonstrated how the design decisions for the action space impacted learning and performance. There is a chance the existing solutions could be improved further with small adjustments to the environment design. In \cite{hamilton2024investigating}, the authors limit their investigations to changes in the action space. This work builds off their learned lessons to explore how changes to the observation space, i.e. the input to the learned controller, impact the training and performance. 

This work investigates changes to the observation space for the inspection environment introduced in \cite{vanWijkAAS_23} in two groupings. The first group looks at the use of sensors, which provide additional information about the environment outside of the position and velocity information. The sensors were designed to aid in completing the inspection task, but this study looks to validate that they are a help and not a hindrance. The second group looks into the affects of the agent's view through changing the reference frame. The inspection environment was designed with everything centered around the spacecraft under inspection. This view point has the RL agent operating in a 3rd-person field of view, similar to a human driving an RC car through a racetrack and watching from a fixed position. This is in contrast to traditional RL formulations where the RL agent controls the system from a 1st-person point of view. 

These groups of investigations seek to answer the following questions:
\begin{enumerate}
    \item Are RL agents capable of learning a successful policy without the added sensors?
    \item Are there any sensors that hinder learning?
    \item Does changing the reference frame from chief-centered to agent-centered have an impact?
\end{enumerate}

\section{Background} \label{sec:background}

\subsection{Deep Reinforcement Learning} \label{sec:rl}

\begin{figure}[t]
    \centering
    \includegraphics[width=.8\columnwidth]{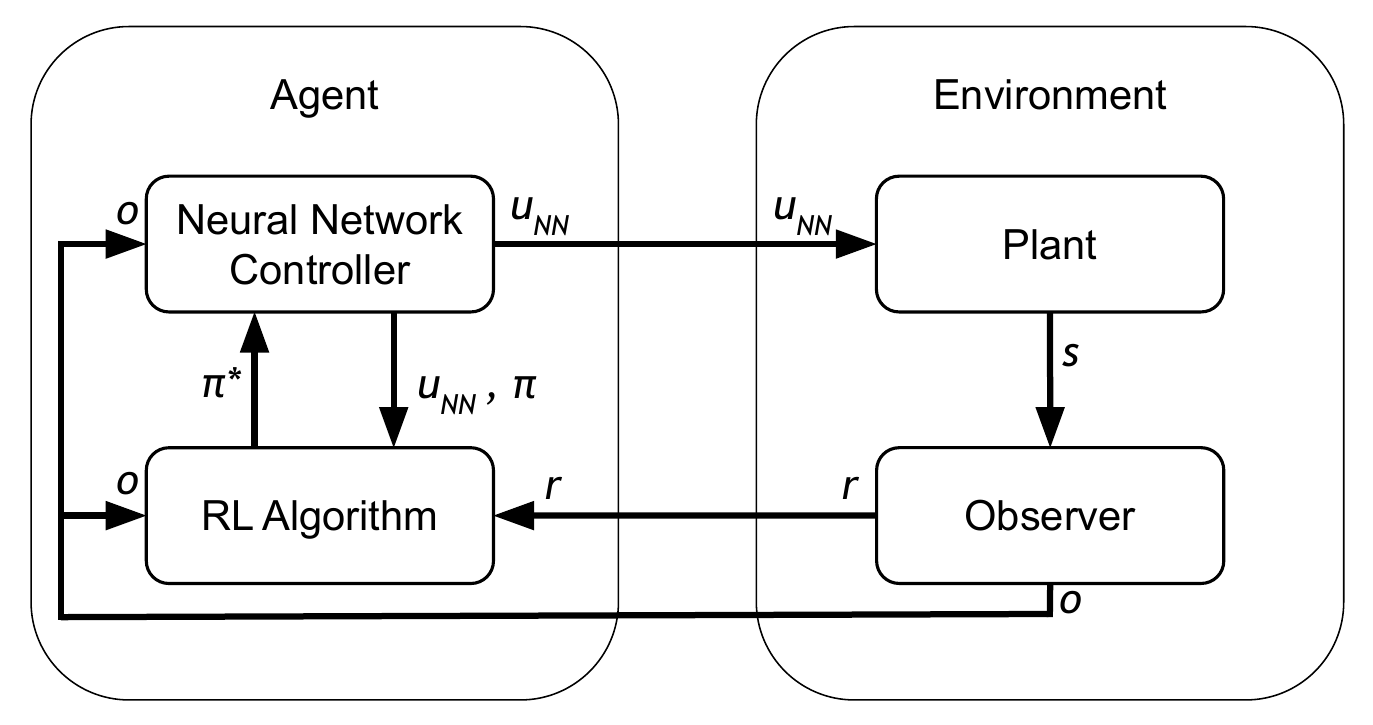}
    \caption{DRL training loop from \cite{hamilton2023ablation}.}
    \label{fig:rta_off}
\end{figure}

\textit{Reinforcement Learning} (RL) is a subset of machine learning where learning agents act in an environment, learn through experience, and improve their performance according to a reward function. \textit{Deep Reinforcement Learning} (DRL) is a newer branch of RL in which a neural network is used to approximate the behavior function, i.e. policy $\policy$.
The basic construction of the DRL approach is shown in \figref{fig:rta_off}. 
The agent uses a \textit{Neural Network Controller} (NNC) trained by the RL algorithm to take actions in the environment, which represent any dynamical system, from Atari simulations (\cite{hamilton2020sonic, alshiekh2018safe}) to complex robotics scenarios (\cite{brockman2016gym, fisac2018general, mania2018simple, jang2019simulation, hamilton2022zero}). 

Reinforcement learning is based on the \textit{reward hypothesis} that all goals can be described by the maximization of expected return, i.e. the cumulative reward \cite{silver2015}. During training, the agent uses the input observation, $\obs$, to choose an action, $\action_{NN}$. The action is then executed in the environment, and the state, $\state$, transitions according to the plant dynamics. The updated state is assigned a scalar reward, $\reward$, and transformed into the next observation vector. The singular loop of executing an action and receiving a reward and next observation is referred to as a \textit{timestep} and is used as a measure of how long an agent has been training. Relevant values, like the input observation, action, and reward are collected as a data tuple, i.e. \textit{sample}, by the RL algorithm to update the current NNC policy, $\policy$, to an improved policy, $\policy^*$. How often these updates are done is dependent on the RL algorithm.

This work uses the \textit{Proximal Policy Optimization} (PPO) algorithm as the DRL algorithm of choice. PPO has demonstrated success in the space domain for multiple tasks and excels in finding optimal policies across many other domains \cite{schulman2017proximal, hamilton2023ablation, ravaioli2022safe, dunlap2021Safe, vanWijkAAS_23, hamilton2024investigating}.

\subsection{Spacecraft Inspection Environment} \label{sec:environment}

\begin{figure}[t]
    \centering
    \includegraphics[width=.8\columnwidth]{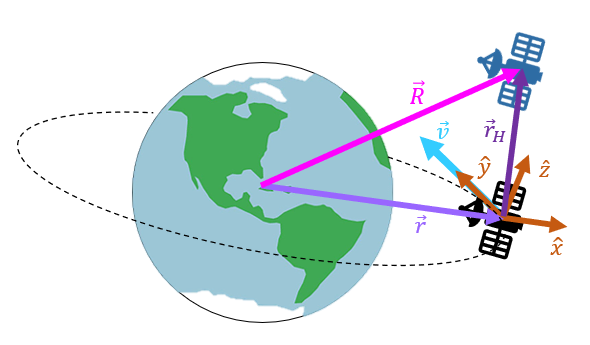}
    \caption{Deputy spacecraft navigating around a chief spacecraft in Hill's Frame from \cite{dunlap2024run}.}
    \label{fig:Hills}
\end{figure}

This work focuses on the \textit{inspection} task introduced in \cite{vanWijkAAS_23}. The RL agent learns a policy for navigating a deputy spacecraft about a chief spacecraft in order to observe its entire surface. In this case, the chief is modeled as a sphere of $99$ inspectable points distributed uniformly across the surface. The attitude of the deputy is not modeled, because it is assumed that the deputy is always pointed towards the chief. In order for a point to be inspected, it must be within the field of view of the deputy (not obstructed by the near side of the sphere) and illuminated by the Sun. Illumination is determined using a binary ray tracing technique, where the Sun appears to rotate in the $\hat{x}-\hat{y}$ plane in Hill's frame at the same rate as mean motion of the chief's orbit, $\meanmotion$.

The dynamics are modeled using the Clohessy-Wiltshire equations \cite{clohessy1960terminal} in Hill's frame \cite{hill1878researches}, which is a linearized relative motion reference frame centered around the chief spacecraft, which is in a circular orbit around the Earth. As shown in \figref{fig:Hills}, the origin of Hill's frame, $\mathcal{O}_H$, is located at the chief's center of mass. The unit vector $\hat{x}$ points in the radial direction away from the center of the Earth. The unit vector $\hat{y}$ points in the in-track direction, which point in the direction of chief's motion. And the unit vector $\hat{z}$ is normal to $\hat{x}$ and $\hat{y}$, which is the cross-track direction. The relative motion dynamics between the deputy and chief are, 
\begin{equation} \label{eq: system dynamics}
    \dot{\state} = A {\state} + B\action,
\end{equation}
where $\state$ is the state vector $\state=[x,y,z,\dot{x},\dot{y},\dot{z}]^T \in \mathbb{R}^6$, $\action$ is the control vector, i.e. action, and,
\begin{align}
\centering
    A = 
\begin{bmatrix} 
0 & 0 & 0 & 1 & 0 & 0 \\
0 & 0 & 0 & 0 & 1 & 0 \\
0 & 0 & 0 & 0 & 0 & 1 \\
3\meanmotion^2 & 0 & 0 & 0 & 2\meanmotion & 0 \\
0 & 0 & 0 & -2\meanmotion & 0 & 0 \\
0 & 0 & -\meanmotion^2 & 0 & 0 & 0 \\
\end{bmatrix}, 
    B = 
\begin{bmatrix} 
 0 & 0 & 0 \\
 0 & 0 & 0 \\
 0 & 0 & 0 \\
\frac{1}{\mass} & 0 & 0 \\
0 & \frac{1}{\mass} & 0 \\
0 & 0 & \frac{1}{\mass} \\
\end{bmatrix}.
\end{align}
Here, $\meanmotion = 0.001027$rad/s is the mean motion of the chief's orbit, and $\mass = 12$kg is the mass of the deputy. The control vector, $\action= [F_x,F_y,F_z]^T$, applies forces exerted by thrusters along each axis. Each force value is either -0.1, 0.0, or 0.1 as determined by the agent. These discrete option are the ones that worked the best in \cite{hamilton2024investigating}. Both spacecraft are modeled as point masses.

While the main objective of the task is to inspect all points, a secondary objective is to minimize fuel use. This is considered in terms of $\deltav$, where,
\begin{equation}
    \deltav = \frac{|F_{x}| + |F_{y}| + |F_{z}|}{m} \deltat.
\end{equation}
For this task, $\deltat = 10$ seconds.

\subsubsection{Initial and Terminal Conditions}

Each episode is randomly initialized given the following parameters. First, the Sun is initialized at a random angle with respect to the $\hat{x}$ axis so that $\sunangle \in [0, 2\pi]$rad. Next, the deputy's position is sampled from a uniform distribution for the parameters: radius $\radius \in [50, 100]$m, azimuth angle $\azimuth \in [0, 2\pi]$rad, and elevation angle $\elevationangle \in [-\pi/2, \pi/2]$rad. The position is then computed as,
\begin{equation} \label{eq:init}
    \begin{gathered}
        x = \radius \cos(\azimuth) \cos(\elevationangle), \\
        y = \radius \sin(\azimuth) \cos(\elevationangle), \\
        z = \radius \sin(\elevationangle). \\
    \end{gathered}
\end{equation}
If the deputy's initialized position results in pointing within 30 degrees of the Sun, the position is negated, which directs the deputy to point away from the Sun and towards illuminated points. This prevents unsafe and unrealistic initialization, as sensors can burnout when pointed directly at the Sun. Finally, the deputy's velocity is similarly sampled from a velocity magnitude $\totalvel \in [0, 0.3]$m/s, azimuth angle $\azimuth \in [0, 2\pi]$rad, and elevation angle $\elevationangle \in [-\pi/2, \pi/2]$rad, and the velocity is computed using the same technique as \eqref{eq:init}.

An episode is terminated under the following conditions: 
\begin{enumerate}
    \item the deputy inspects all $99$ points, 
    \item the deputy crashes into the chief (enters within a minimum relative distance of $15$m, where the chief and deputy have radii $10$ and $5$m respectively), 
    \item the deputy exceeds a maximum relative distance from the chief of $800$m, and/or 
    \item the simulation exceeds $1223$ timesteps (the time for the Sun to appear to orbit the chief twice, or $3.4$hrs).
\end{enumerate}

\subsubsection{Original Observations}

The environment is partially observable, using sensors to condense full state information into manageable components of the observation space. At each timestep, the agent receives an observation comprised of the following components. The first component is the deputy's current position in Hill's frame, where each element is divided by a value of $100$ to ensure most values fall in the range $[-1, 1]$. The second component is the deputy's current velocity in Hill's frame, where each element is multiplied by a value of $2$ to ensure most values fall in the range $[-1, 1]$. The third component is the angle describing the Sun's position with respect to the $\hat{x}$ axis, $\sunangle$. The fourth component is the total number of points that have been inspected so far during the episode, $\numpoints$, divided by a value of $100$. The final component is a unit vector pointing towards the nearest cluster of uninspected points, where clusters are determined using k-means clustering. The resulting observation is $\obs = [x, y, z, \dot{x}, \dot{y}, \dot{z}, \numpoints, \sunangle, x_{UPS}, y_{UPS}, z_{UPS}]$. This representation is the observation unless otherwise specified, because the focus of this work is on changing what is included in the observation and how it is represented.

\subsubsection{Reward Function}

The reward function consists of the following three elements\footnote{The reward function was defined in \cite{vanWijkAAS_23}, and the authors determined that the specified configuration produces the desired behavior. An exploration of the reward function is outside the scope of this work.}. First, a reward of $+0.1$ is given for every new point that the deputy inspects at each timestep. Second, a negative reward is given that is proportional to the $\deltav$ used at each timestep. This is given as $-w*\deltav$, where $w$ is a scalar multiplier that changes during training to help the agent first learn to inspect all points and then minimize fuel usage. At the beginning of training, $w=0.001$. If the mean percentage of inspected points for the previous training iteration exceeds 90\%, $w$ is increased by $0.00005$, and if this percentage drops below 80\% for the previous iteration, $w$ is decreased by the same amount. $w$ is enforced to always be in the range $[0.001, 0.1]$. Finally, a reward of $-1$ is given if the deputy collides with the chief and ends the episode. This is the only sparse reward given to the agent.
For evaluation, a constant value of $w=0.1$ is used, while all other rewards remain the same.

\section{General Experiment Setup} \label{sec:setup}

The experiments in this work are split into two groups. The first group of configurations focuses on modifying the observation space by adding and removing sensors, which provide additional information about how much the agent has inspected so far. The second group of configurations focuses on modifying the reference frame used to determine position information in the agents' observation. 

This work uses ablation studies to single out and identify how each component/modification to the observation space impacts the training and performance of the RL agents. Ablation studies come from the neuroscience community where they were used to identify the purpose of different parts of the rat brain by removing each part and observing the effects. This work employs a similar process by singling out and removing the individual components/modifications to observe their impact. To this end, RL agents are trained for each of these unique configurations with all hyperparameters, listed in \tabref{tab:hparams}, held constant to observe the impact of only the change in configuration. Additionally, the RL training is repeated across 10 random seeds to ensure the differences are repeatable and not only the result of a favorable set of training conditions \cite{henderson2018deep, mania2018simple, agarwal2021deep, hamilton2023ablation}.

\begin{table}[]
    \centering
    \caption{PPO Hyperparameters Used for All Training}
    \label{tab:hparams}
    \begin{tabular}{lc} \hline
        Parameter                       & Value \\ \hline
        Number of SGD iterations        & 30 \\ 
        Discount factor $\gamma$        & 0.99 \\
        GAE-$\lambda$                   & 0.928544 \\
        Max episode length              & 1223 \\
        Rollout fragment length         & 1500 \\
        Train batch size                & 1500 \\
        SGD minibatch size              & 1500 \\
        Total timesteps                 & $5 \times 10^6$ \\
        Learning rate                   & $5 \times 10^{-5}$ \\
        KL initial coefficient          & 0.2 \\
        KL target value                 & 0.01 \\
        Value function loss coefficient & 1.0 \\
        Entropy coefficient             & 0.0 \\
        Clip parameter                  & 0.3 \\
        Value function clip parameter   & 10.0 \\
        \hline
    \end{tabular}
\end{table}

The metrics of interest are Total Reward, Inspected Points, Episode Length, Success Rate, and Delta V. Comparisons between configurations are measured using the \textit{InterQuartile Mean} (IQM)\footnote{The IQM removes the top and bottom $25\%$ of results to remove outliers and compute the mean on the middle $50\%$.}, and the $95\%$ confidence interval (CI) around it because it represents a better prediction of future studies as it is not unduly affected by outliers and has a smaller uncertainty even with a handful of runs \cite{agarwal2021deep}. 

Impact is measured by observing changes to the agents' sample complexity and final performance. Sample complexity is a measure of how quickly agents learn a task. Improved sample complexity will show higher reward, number of inspected points, and/or success after fewer timesteps (in the plots, the curves will be closer to the top left corner). Improved sample complexity for fuel efficiency, i.e. lower Delta V, will be lower with fewer timesteps (in the plots, the curves will be closer to the bottom left corner).

Final performance is measured using the final saved policies evaluated for 100 episodes to record the expected behavior of the policy if it were used for the desired task. This provides more insight into the final product, ignoring differences in the training process, only concerned with the ultimate performance of the models once trained. Final performance comparisons are more 1:1 since training time is no longer considered. The most fuel-efficient configuration will have the smallest measured Delta V; the best performing configuration will have the highest total reward; and the most successful configuration will have the highest success rate.

\section{Investigating Sensor Selection} \label{sec:sensors}

This experiment was designed to identify which sensor inputs are essential for agents to successfully learn to complete the task. While developing the environment, sensors were added in an effort to help agents learn the task since minimal improvement was seen using only the state values (i.e. position and velocity). Throughout the efforts to find a good baseline, new sensors were added incrementally, but never removed. In this experiment, the sensors are isolated to see just how much they are impacting the training and performance of the agents. The results focus on identifying which sensors are necessary for completing the task, and which sensors help speed up training.

Providing the agent with the full state observability, i.e. providing an input for each point and whether it has been inspected, would be the most straight-forward approach for solving this issue. However, this makes the network substantially larger, which would likely require more training time, and will not extend to new scenarios where more points are needed to cover the chief, or the points are spread across the chief in a different pattern.

To this end, additional sensors were developed to include in the observation space with the intent of helping provide the agent with enough information to decide which directions to travel to inspect all the points. These are the \textit{Count Sensor}, \textit{Sun Angle Sensor}, and the \textit{Uninspected Points Sensor}.

The \textit{Count Sensor} provides the agent with an integer giving the number of points inspected so far in the episode. With each step in the environment, it updates to reflect the additional points inspected.

The \textit{Sun Angle Sensor} provides the agent with the angular position of the Sun represented as a single value. In the inspection environment, the Sun rotates clockwise in the $\hat{x}-\hat{y}$ plane of Hill's frame centered on the chief spacecraft. Using this knowledge, the sun angle is measured from the positive $x$-axis to the unit vector pointing from the chief to the sun. The angle is measured positively in the clockwise direction.

The \textit{Uninspected Points Sensor} (UPS) provides the agent with a 3D unit vector pointing towards the nearest cluster of uninspected points. A K-means clustering algorithm is used to identify the clusters of uninspected points. This sensor activates every time new points are inspected or new points are illuminated, scanning for a new largest cluster of uninspected points. The clusters are initialized from the previously identified clusters and the total number of clusters is never more than $num\_inspected\_points / 10$. This sensor helps the agent identify the best direction to move in order to inspect new points. It was also designed for potential re-purposing in the future for a more generalized "region of interest" sensor guiding the agent towards specific regions requiring more thorough inspection.

To this end, agents were trained using 8 different configurations covering all combinations of no sensors, 1 sensor, 2 sensors, and the baseline with all 3 sensors. Nothing is changed in these configurations except for the sensors included in the observation space.

\subsection{Sensors Necessary for Completing the Task}

\begin{figure}[htbp]
    \centering
    \subfigure[]{\includegraphics[width=\linewidth]{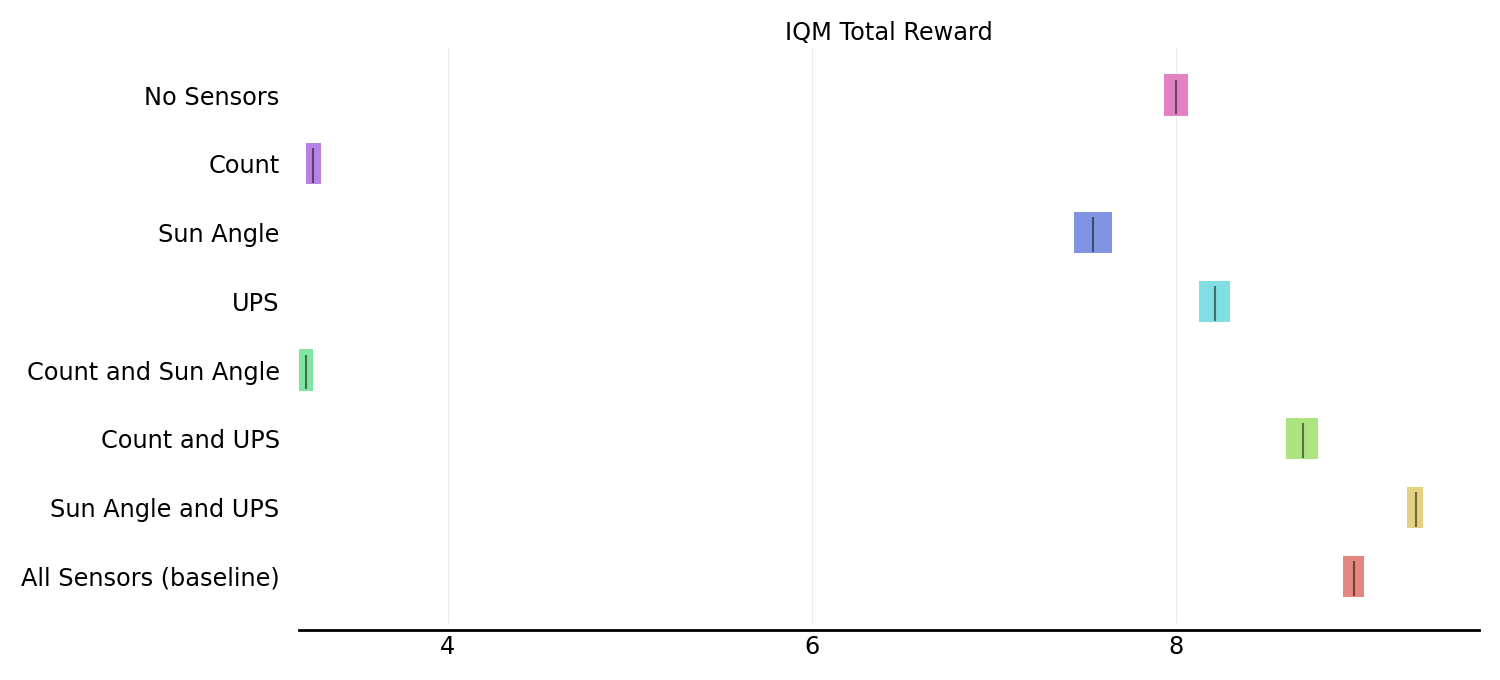}} \\
    \subfigure[]{\includegraphics[width=.8\linewidth]{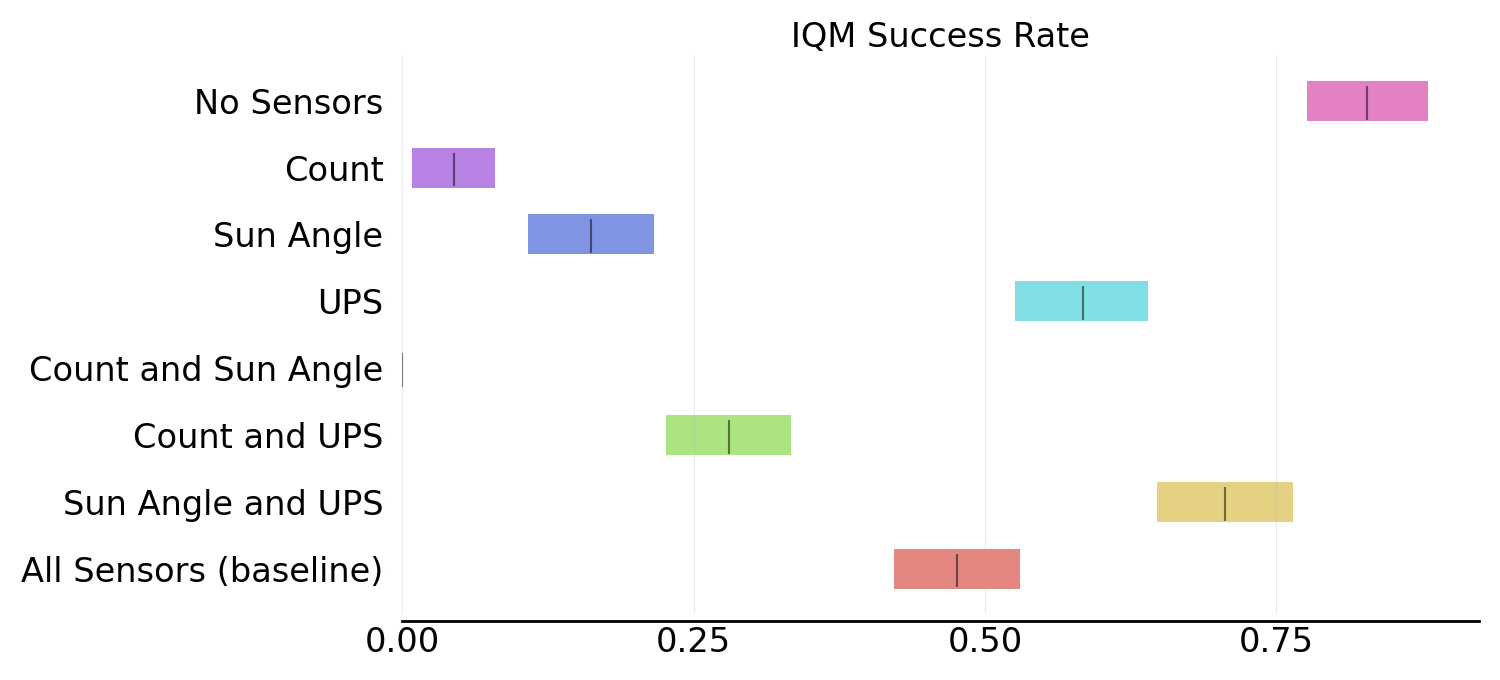}} \quad
    \caption{Final policy performance plots showing the expected level of performance of the final trained models measured in total reward and success rate. The line represents the interquartile mean and the shaded region is the 95\% confidence interval.}
    \label{fig:sensor_final_policy_r_s}
\end{figure}

\begin{figure}[htbp]
    \centering
    \subfigure[]{\includegraphics[width=.8\linewidth]{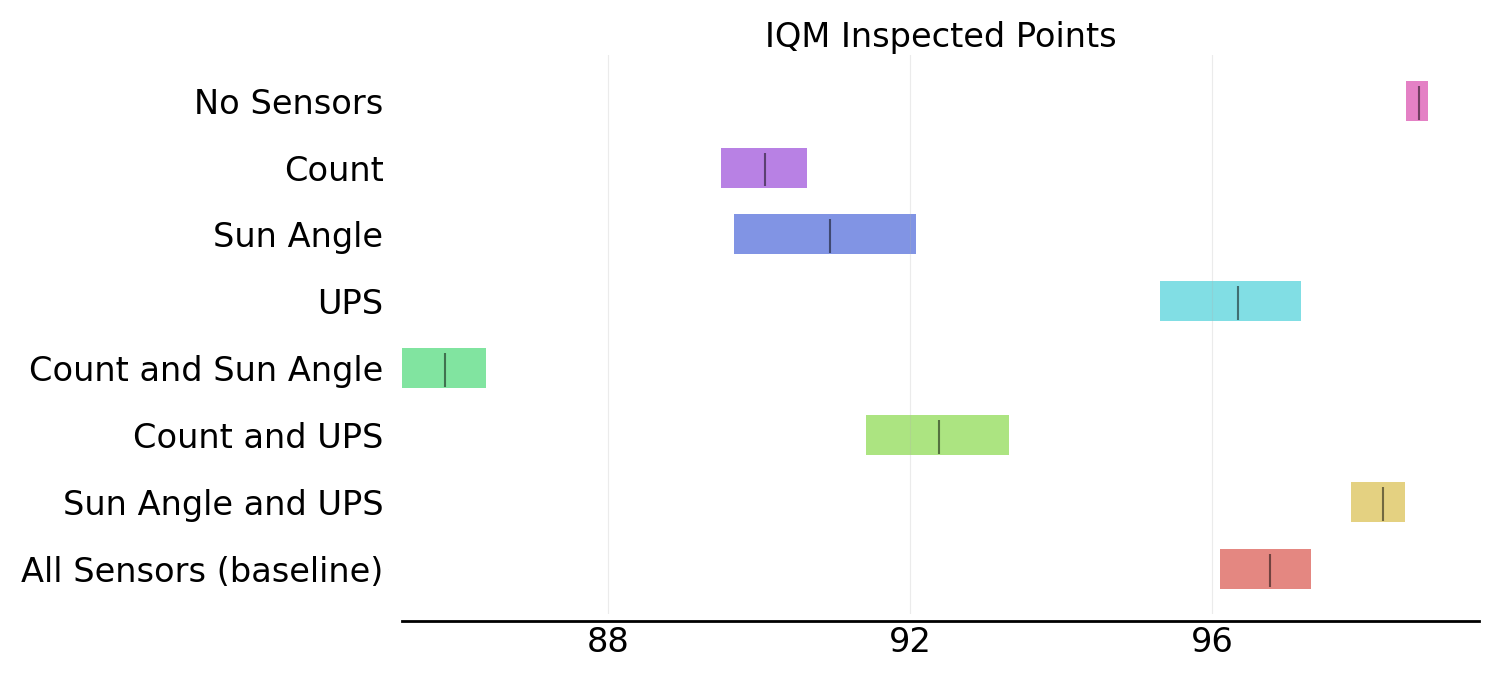}} \\
    \subfigure[]{\includegraphics[width=.8\linewidth]{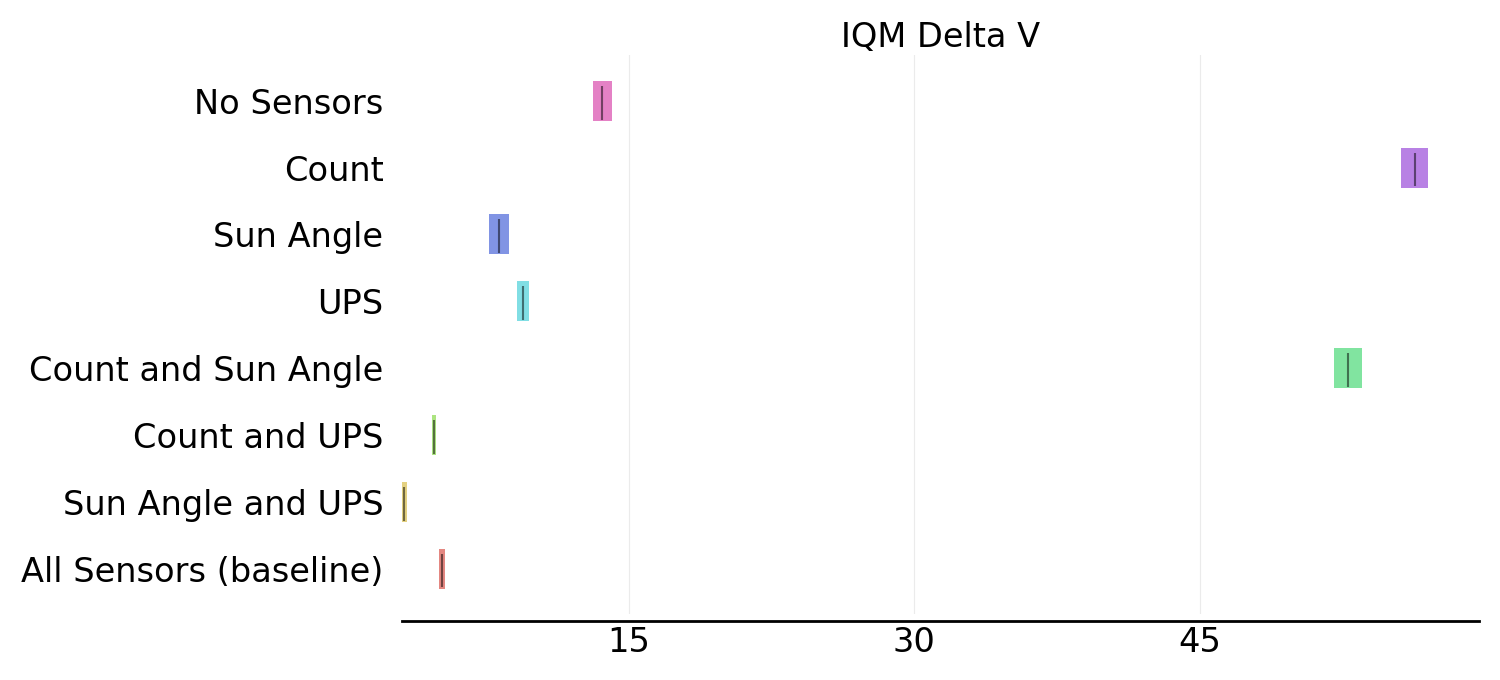}} \\
    \subfigure[]{\includegraphics[width=.8\linewidth]{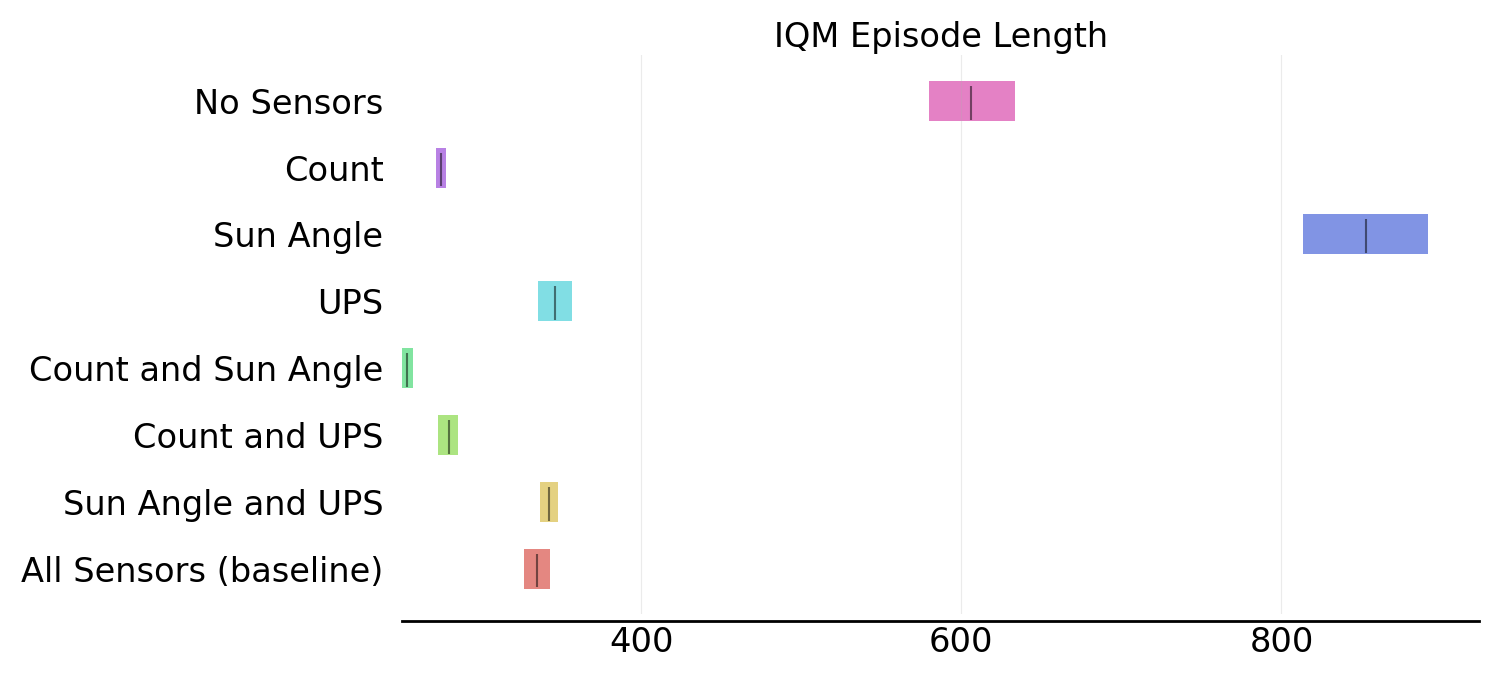}} \\
    \caption{Final policy performance plots showing the expected level of performance of the final trained models measured in number of inspected points, fuel usage, and episode length. The line represents the interquartile mean and the shaded region is the 95\% confidence interval.}
    \label{fig:sensor_final_policy_n_dv_l}
\end{figure}

\begin{table}[htbp]
\centering
\caption{Table of InterQuartile Means and 95\% CI for Sensor Selection Study}
\label{tab:sensors}
\resizebox{1.0\linewidth}{!}{
\begin{tabular}{llllll}
\toprule
Metric Labels &       Total Reward &      Inspected Points &           Episode Length $(s)$ &       Success Rate &               Delta V $(m/s)$ \\
\midrule
No Sensors             &   8.0 [7.93, 8.07] &  98.74 [98.56, 98.86] &  605.99 [579.76, 633.78] &  0.83 [0.78, 0.88] &  13.62 [13.14, 14.11]  \\
Count                  &   3.26 [3.22, 3.3] &   90.08 [89.5, 90.63] &  275.16 [272.03, 278.25] &  0.04 [0.01, 0.08] &  56.29 [55.59, 56.98]  \\
Sun Angle              &  7.55 [7.44, 7.65] &  90.94 [89.67, 92.08] &  852.88 [813.19, 891.35] &  0.16 [0.11, 0.22] &      8.2 [7.69, 8.73]  \\
UPS                    &  8.21 [8.13, 8.29] &  96.34 [95.31, 97.18] &  346.37 [335.58, 356.83] &  0.58 [0.53, 0.64] &     9.46 [9.15, 9.79]  \\
Count and Sun Angle    &  3.22 [3.18, 3.26] &  85.84 [85.28, 86.39] &  254.04 [250.81, 257.29] &     0.0 [0.0, 0.0] &   52.8 [52.04, 53.54]  \\
Count and UPS          &   8.7 [8.61, 8.78] &  92.38 [91.41, 93.31] &   279.73 [273.4, 285.84] &  0.28 [0.23, 0.33] &     4.78 [4.69, 4.89]  \\
Sun Angle and UPS      &  9.32 [9.27, 9.36] &  98.26 [97.84, 98.55] &  342.37 [337.06, 347.92] &  0.71 [0.65, 0.76] &     3.23 [3.11, 3.35] \\
All Sensors (baseline) &  8.98 [8.92, 9.03] &  96.76 [96.11, 97.31] &  334.93 [327.12, 342.79] &  0.48 [0.42, 0.53] &     5.18 [5.03, 5.35] \\
\bottomrule
\end{tabular}
}
\end{table}

The final policy evaluations from training all 8 different configurations are displayed in Figures \ref{fig:sensor_final_policy_r_s} and \ref{fig:sensor_final_policy_n_dv_l}, and the values are written out in \tabref{tab:sensors}. The results highlight how RL agents are capable of learning a successful policy without the aide of any additional sensors, only using position and velocity information to direct motion in a way that inspects all 99 points more than 75\% of the time and only missing one point when unsuccessful. However, the solution is very costly in terms of fuel use, using approximately three times more fuel than the \textit{Sun Angle and UPS} configuration to inspect the 99th inspection point an additional $10\%$ of the time. Therefore, while RL is successful at training a policy to complete the inspection task without additional sensors, the best performing solution comes from the \textit{Sun Angle and UPS} configuration. 

\begin{figure}
    \centering
    \subfigure[No Sensors configuration, seed: 2875]{\includegraphics[width=.86\linewidth]{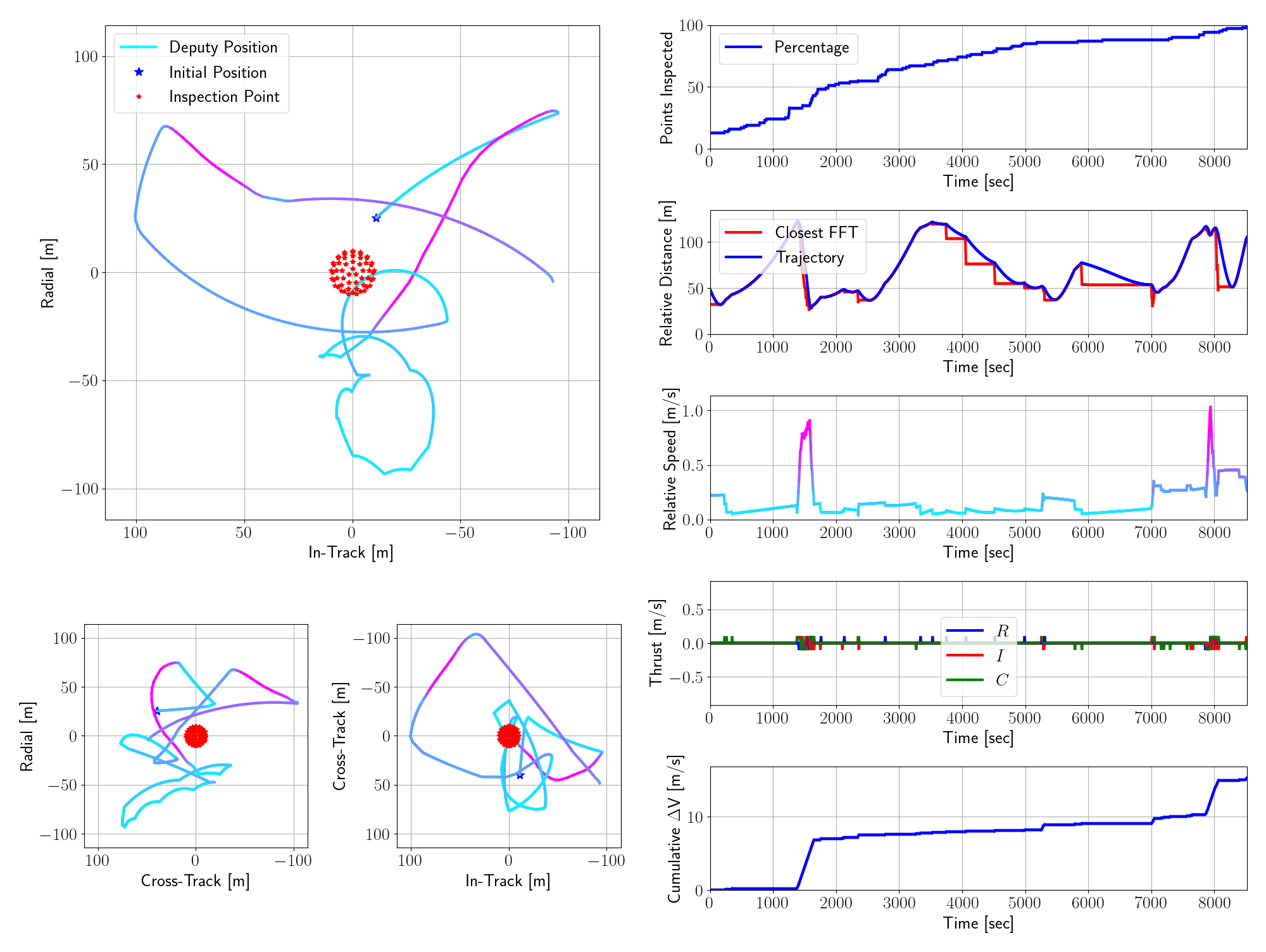}} \\
    \subfigure[No Sensors configuration, seed: 5761]{\includegraphics[width=.86\linewidth]{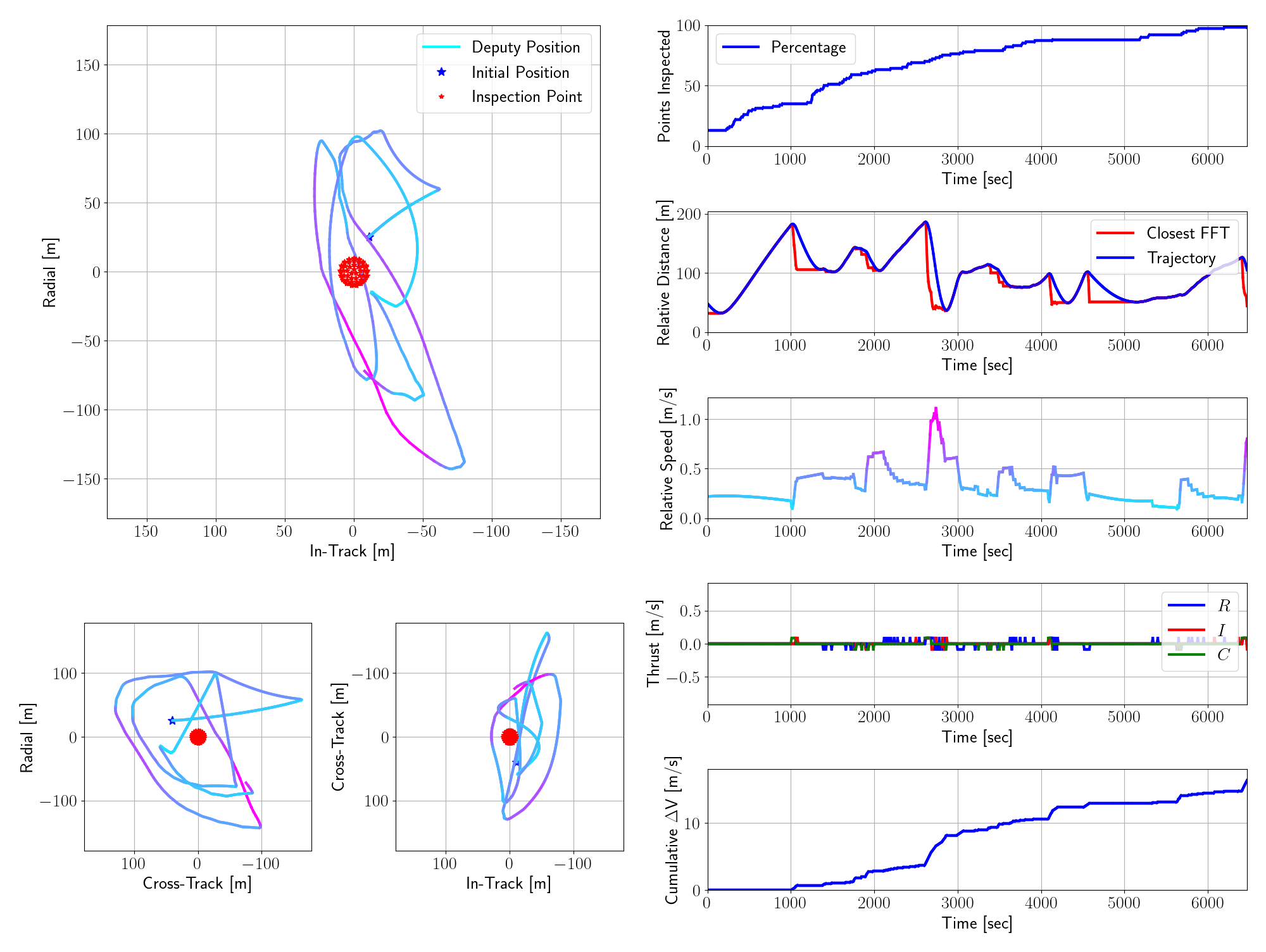}} 
    \caption{Example episodes using policies trained with the \textit{No Sensors} configuration.}
    \label{fig:no_sensors_episode_comparisons}
\end{figure}

\begin{figure}
    \centering
    \subfigure[Sun Angle and UPS configuration, seed: 2875]{\includegraphics[width=.86\linewidth]{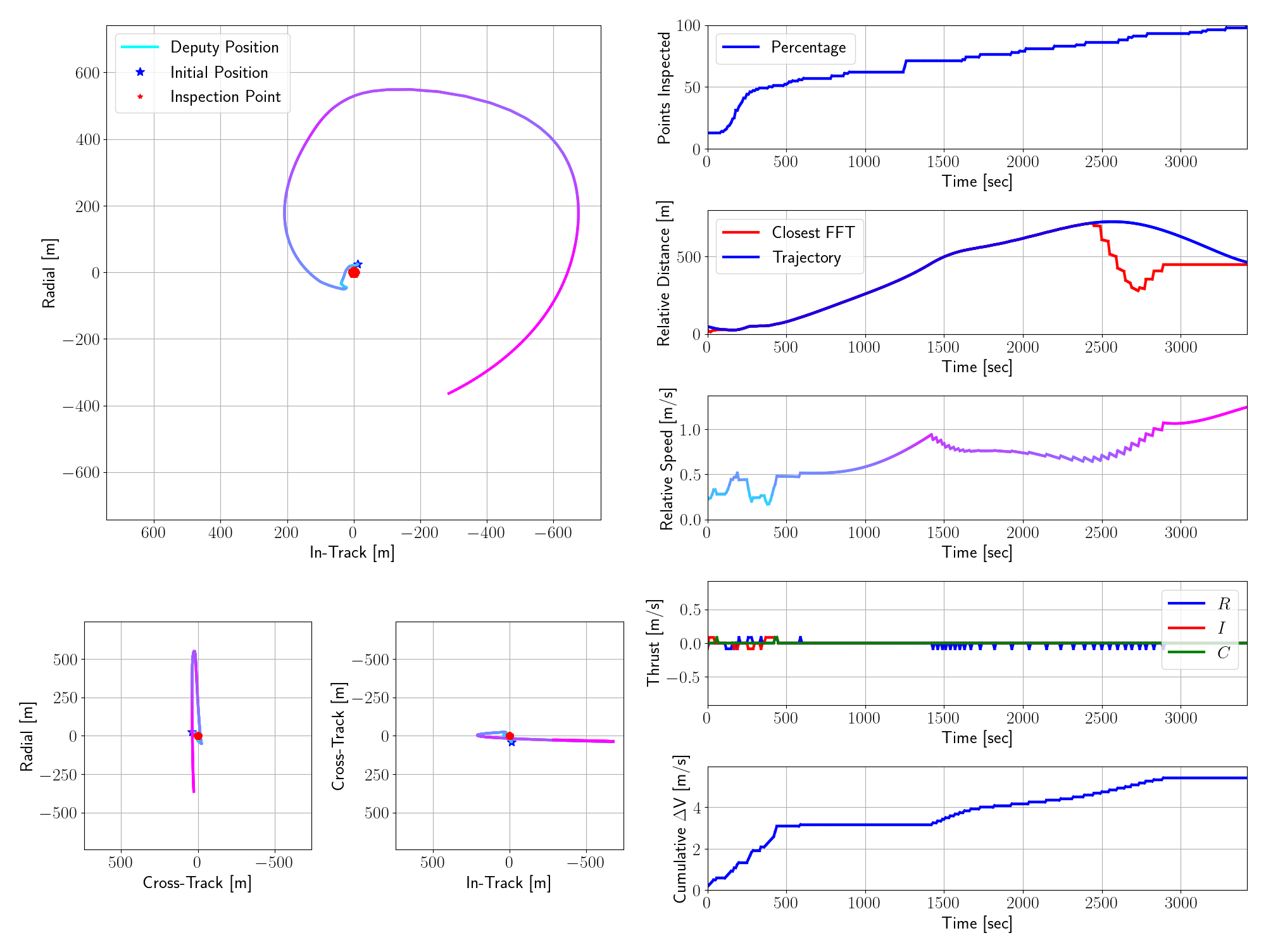}} \\
    \subfigure[Sun Angle and UPS configuration, seed: 5761]{\includegraphics[width=.86\linewidth]{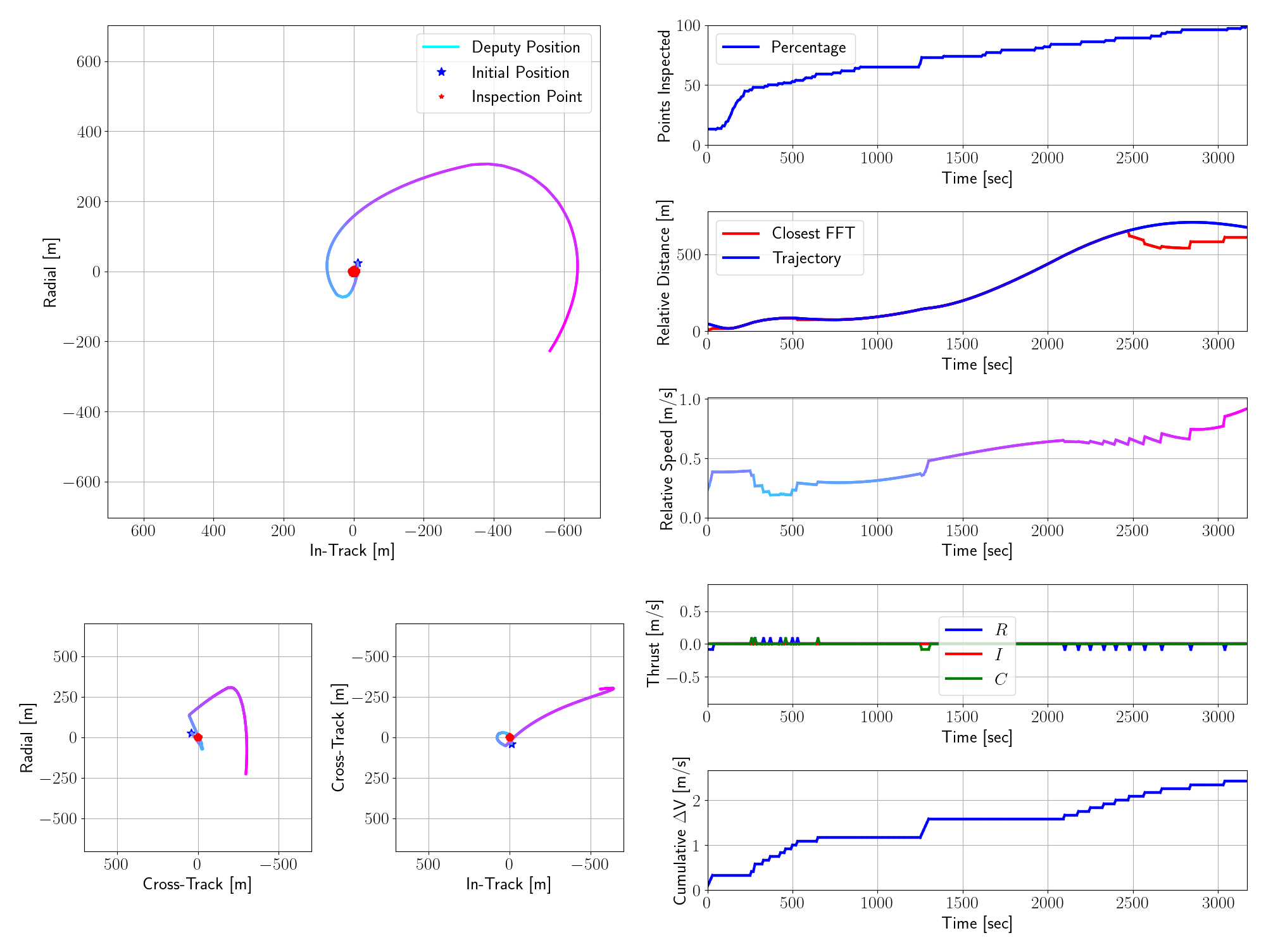}}
    \caption{Example episodes using policies trained with the \textit{Sun Angle and UPS} configuration.}
    \label{fig:saups_episode_comparisons}
\end{figure}

Furthermore, while the RL agents were successful at learning to complete the inspection task without the aide of any additional sensors, the learned policies had more variations in behaviors. Because each configuration was trained with 10 different random seeds, differences in the learned behavior can be observed. \figref{fig:no_sensors_episode_comparisons} shows an example of different learned behaviors trained without sensors. While both behaviors inspect all points and use a similar amount of fuel, they have very different flight paths from the same starting point. In contrast, the examples of policies trained with the \textit{Sun Angle and UPS} configuration, shown in \figref{fig:saups_episode_comparisons}, have a more similar flight path moving the deputy outwards and orbiting around the chief in a clockwise motion. The additional information about where the Sun and uninspected points are helps the RL agents converge on more similar optimal behavior.

\subsection{Sensors that Help Agents Train Faster}

\begin{figure}[htbp]
    \centering
    \subfigure[Total Reward]{\includegraphics[width=.3\linewidth]{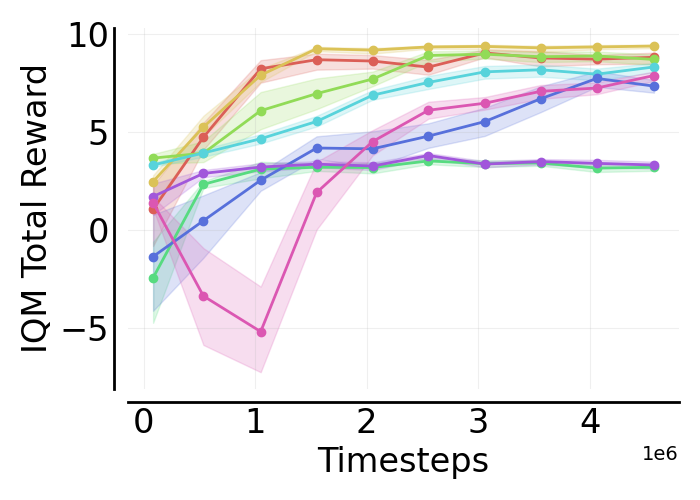}} \quad
    \subfigure[Inspected Points]{\includegraphics[width=.3\linewidth]{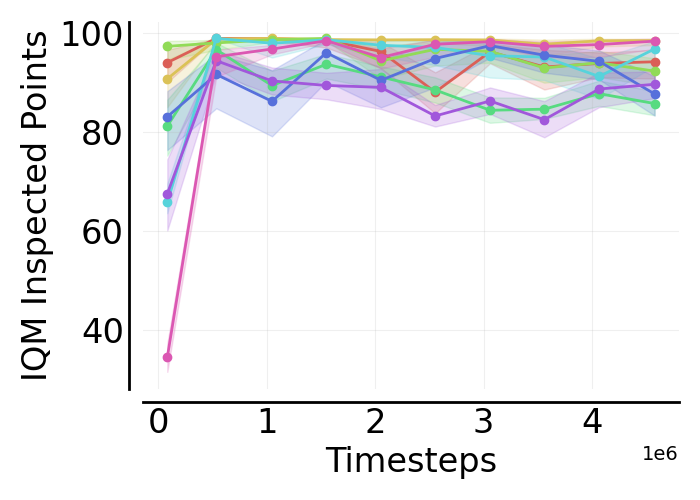}} \quad
    \subfigure[Episode Length]{\includegraphics[width=.3\linewidth]{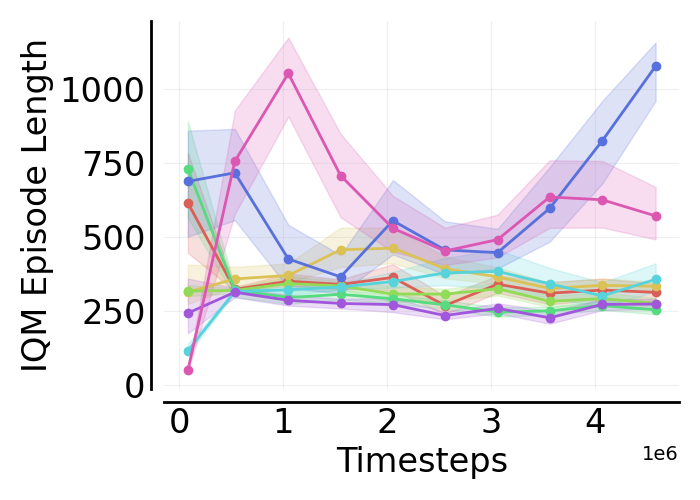}} \\
    \subfigure[Success]{\includegraphics[width=.3\linewidth]{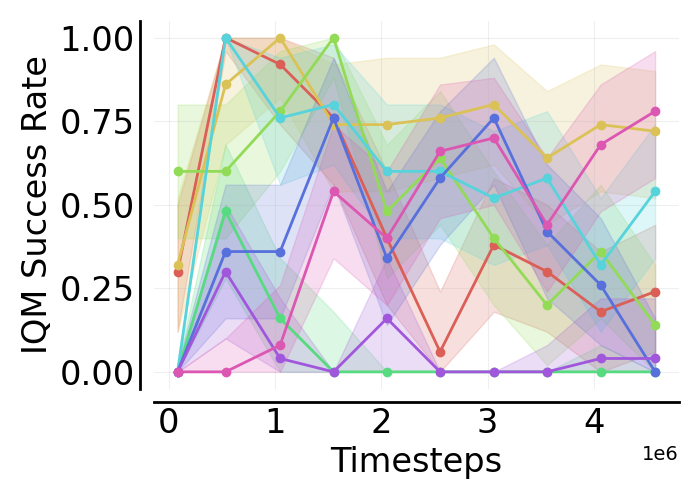}} \quad
    \subfigure[DeltaV]{\includegraphics[width=.3\linewidth]{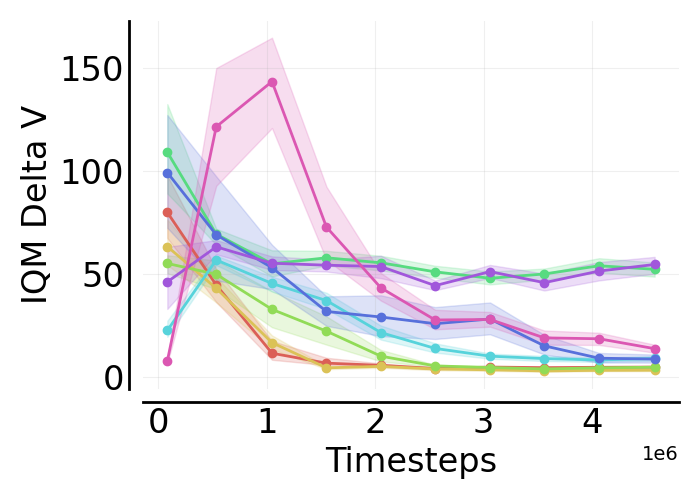}} \quad
    \subfigure[Legend for all configurations]{\includegraphics[width=.3\linewidth]{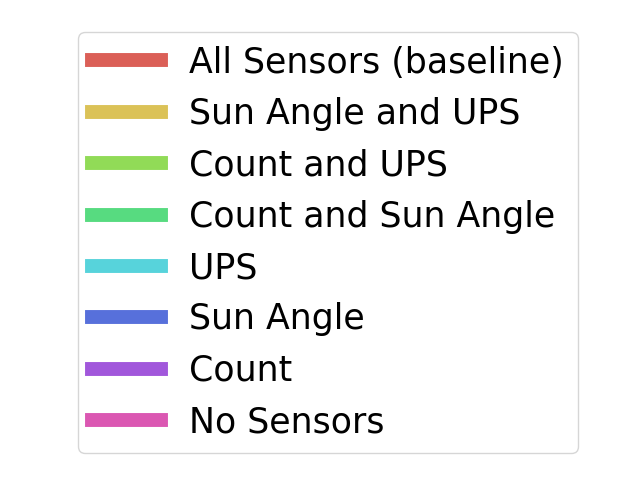}}
    \caption{The sample complexity results for the RL agents trained with different sensors. The line represents the interquartile mean and the shaded region is the 95\% confidence interval.}
    \label{fig:sensors_sample_complexity}
\end{figure}

The sample complexity results in \figref{fig:sensors_sample_complexity} show the uninspected points sensor (UPS) is the most helpful sensor for speeding up training. Generally, all the sensors help speed up training within the first 1 million timesteps. However, the count sensor shows diminishing returns and, in some configurations, prevents further learning by 2 million timesteps. Additionally, \figref{fig:sensor_final_policy} shows the \textit{Count} and \textit{Count and Sun Angle} configurations used more fuel to inspect fewer points than the \textit{No Sensors} configuration, making them detrimental to overall performance. The most effective combination is to include both the sun angle sensor with the UPS. The combination of the two sensors helped improve fuel efficiency faster, significantly speeding up training over the \textit{No Sensors} alternative.  

\section{Investigating Reference Frames} \label{sec:reference_frames}

\begin{figure}
    \centering
    \subfigure[Chief-Centered]{\includegraphics[height=3in]{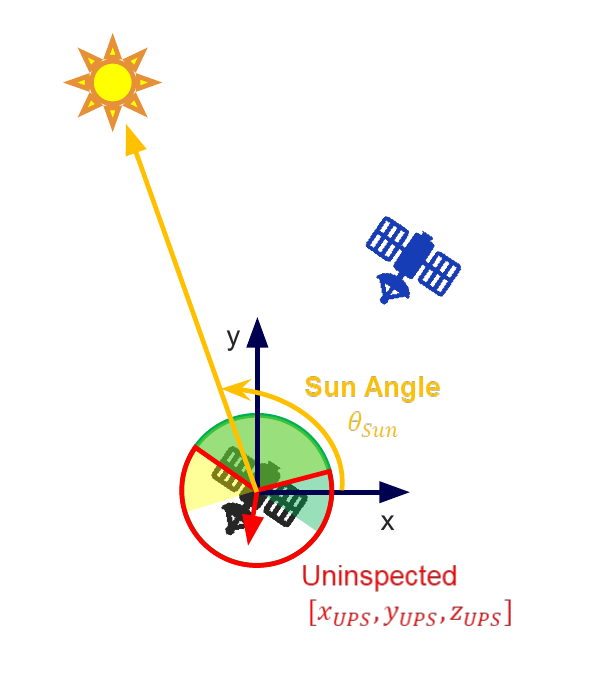}} \quad
    \subfigure[Agent-Centered]{\includegraphics[height=3in]{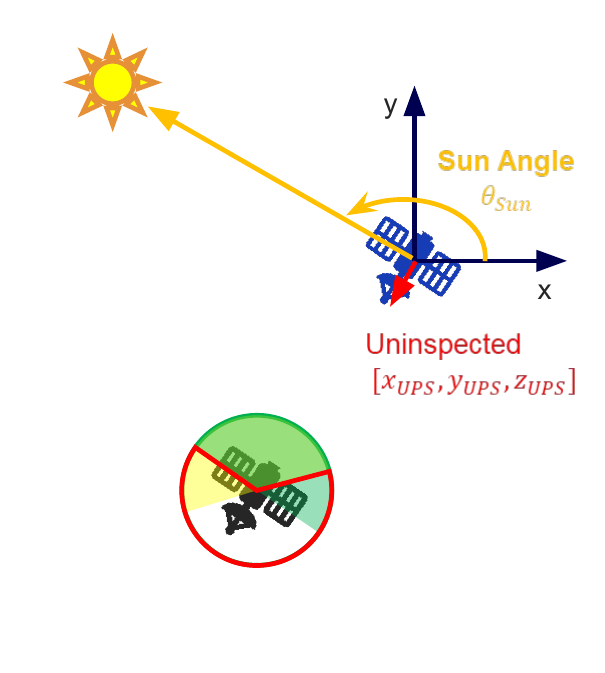}}
    \caption{2-Dimensional representation of how the reference frames change from chief-centered to agent-centered. Note: the distance to the Sun is not to scale.}
    \label{fig:ref-frames}
\end{figure}

This experiment focuses on the impact of the reference frame on learning and performance. The chief-centered reference frame, i.e. Hill's frame, is the default established by the dynamics formulation. Hill's frame treats the chief position as the origin and the deputy, controlled by the RL agent, is informed of its position and velocity in relation to that origin. However, this formulation where the origin is not centered around the agent is not the typical formulation for an RL problem. Instead, an agent-centered reference frame is more like those traditionally used in RL environments, where the agent is at the origin and the observations measure where things are in relation to the agent.\footnote{This can be seen in other navigation tasks like autonomous racing \cite{hamilton2022zero} and safe navigation \cite{ji2023safety}} This requires transforming the observations from Hill's frame to the new origin being the agent/deputy, like shown in \figref{fig:ref-frames}.

For this ablation study, each component that can be altered to the agent-centered reference frame is done incrementally to observe their individual impact. The Position and Velocity components, however, are done together as \textit{Pose} information. Following the lesson learned from the previous study, the configurations in this experiment were trained using the \textit{Sun Angle and UPS} configuration.

\subsection{Converting Chief-Centered, Hill's Frame to the Agent-Centered Frame}

For the pose information, which includes position and velocity, the input is no longer the deputy's position in Hill's frame. Instead, the agent is given the chief's position, relative to the deputy. This is done by inverting the deputy's position. If the deputy is at position $\position_{Hill's} = (x, y, z)$ in Hill's frame, then the chief is at position $\position_{chief} = (-x, -y, -z)$ from the deputy's origin. The same is true for the velocity components.

For the sun angle sensor, no transformation is needed. The Sun is so much further than the maximum distance between chief and deputy, that the largest transformation is negligible.

For the Uninspected Points Sensor (UPS), the transformation requires vector addition to have the vector point from the deputy to the cluster of uninspected points on the chief. The first component is from the deputy to the chief, which is the position of the chief from the deputy's origin, $-\position_{Hill's}$. The second component is the vector from the chief's center to the cluster. The UPS provides a unit vector point from the center of the chief to the center of the largest cluster of uninspected point on it's surface. To bring the vector to the surface, the vector, $\ups_{chief}$, is multiplied by the chief's radius, $\radius$. Adding these two components together creates a vector pointing from the deputy to the uninspected points cluster. The result is:
\begin{equation}
    \ups_{deputy} = \text{unit vector}(\radius(\ups_{chief}) - \position_{Hill's}).
\end{equation}

\subsection{Results for Reference Frames}

\begin{figure}
    \centering
    \subfigure[Total Reward]{\includegraphics[width=.3\linewidth]{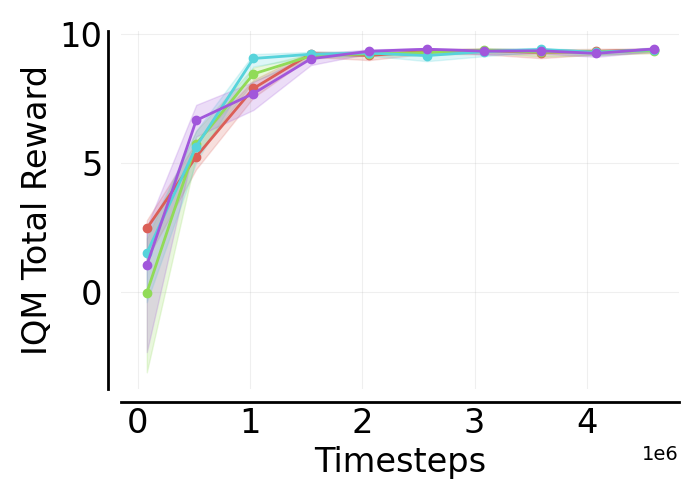}} \quad
    \subfigure[Inspected Points]{\includegraphics[width=.3\linewidth]{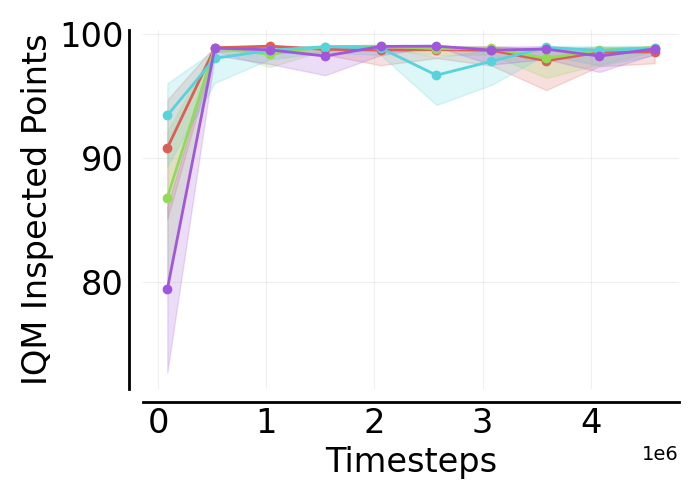}} \quad
    \subfigure[Episode Length]{\includegraphics[width=.3\linewidth]{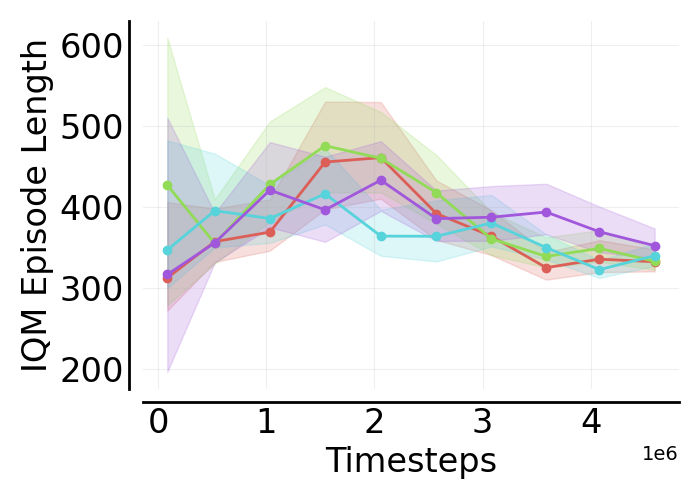}} \\
    \subfigure[Success]{\includegraphics[width=.3\linewidth]{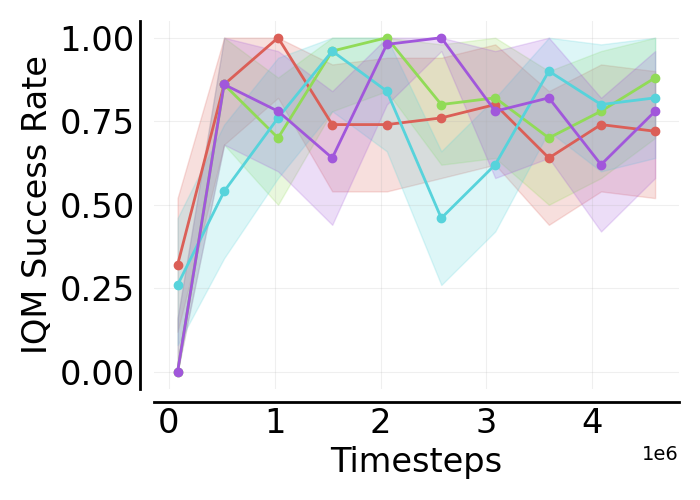}} \quad
    \subfigure[DeltaV]{\includegraphics[width=.3\linewidth]{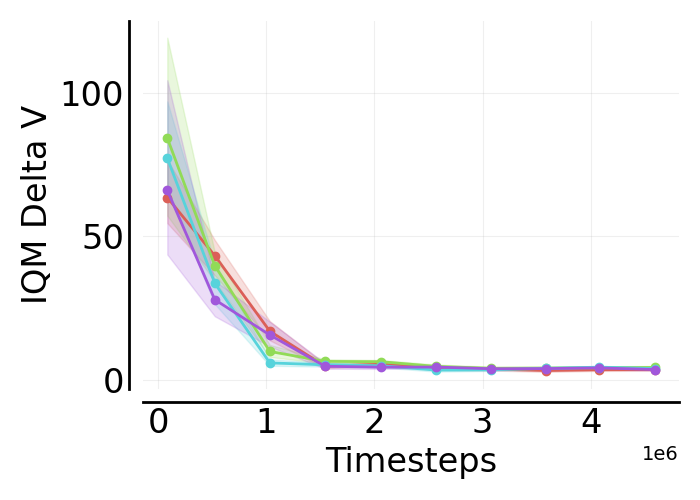}} \quad
    \subfigure[Legend for all configurations]{\includegraphics[width=.3\linewidth]{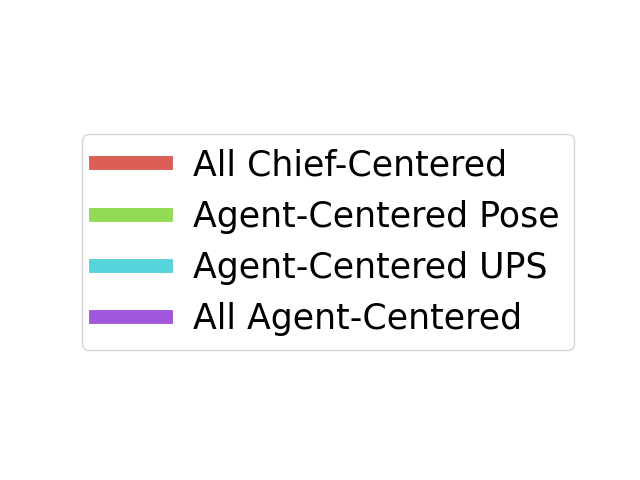}}
    \caption{The sample complexity results for the RL agents trained with different reference frames. The line represents the interquartile mean and the shaded region is the 95\% confidence interval.}
    \label{fig:ref_sample_complexity}
\end{figure}

\begin{figure}[htb]
    \centering
    \subfigure[]{\includegraphics[width=\linewidth]{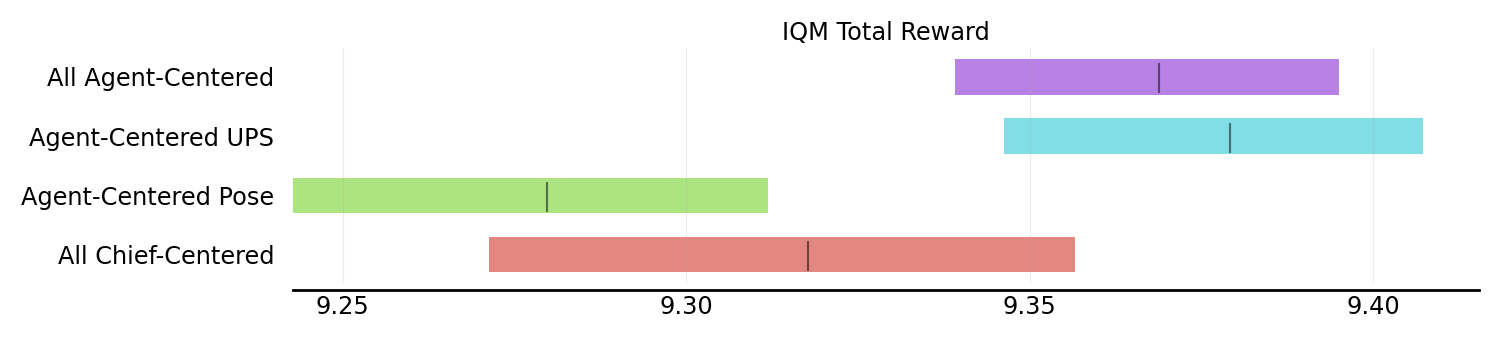}} \\
    \subfigure[]{\includegraphics[width=.8\linewidth]{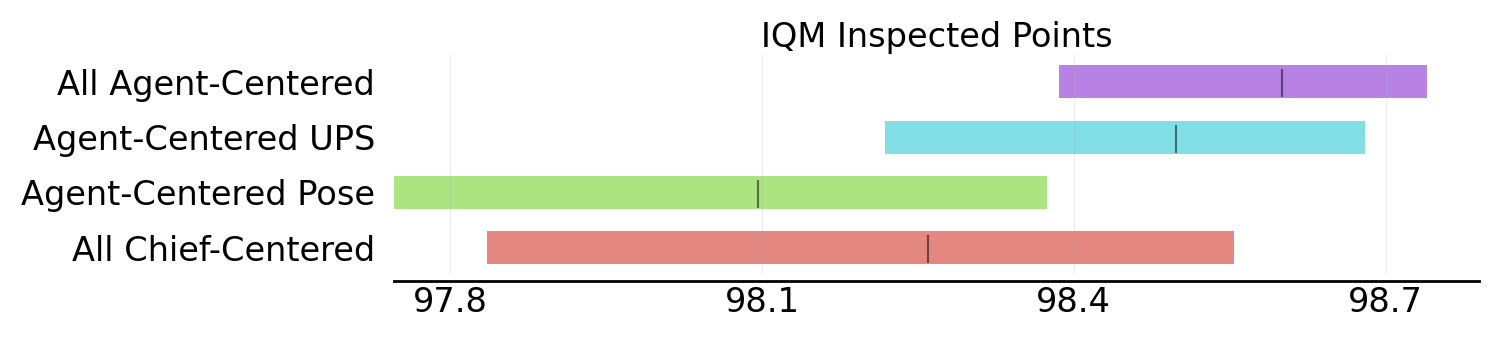}} \quad
    \subfigure[]{\includegraphics[width=.8\linewidth]{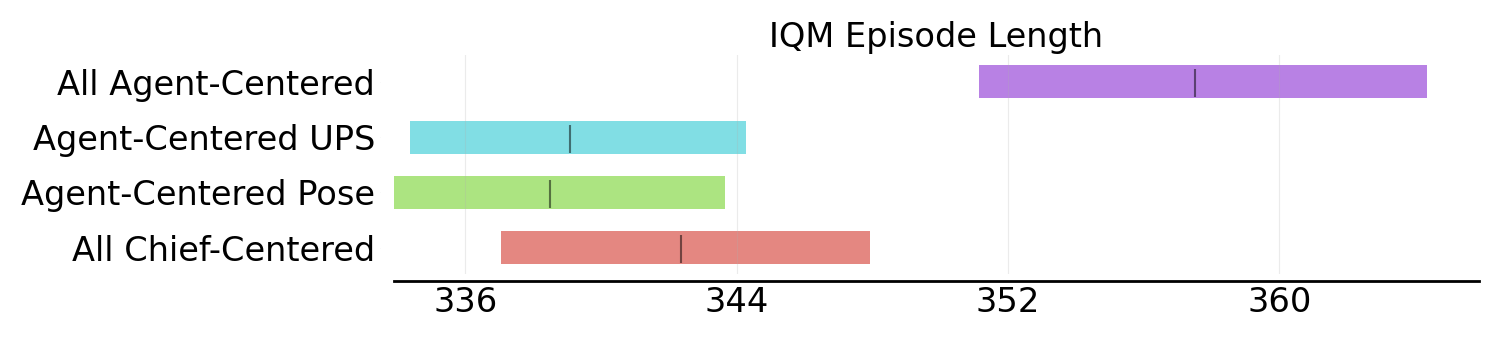}} \\
    \subfigure[]{\includegraphics[width=.8\linewidth]{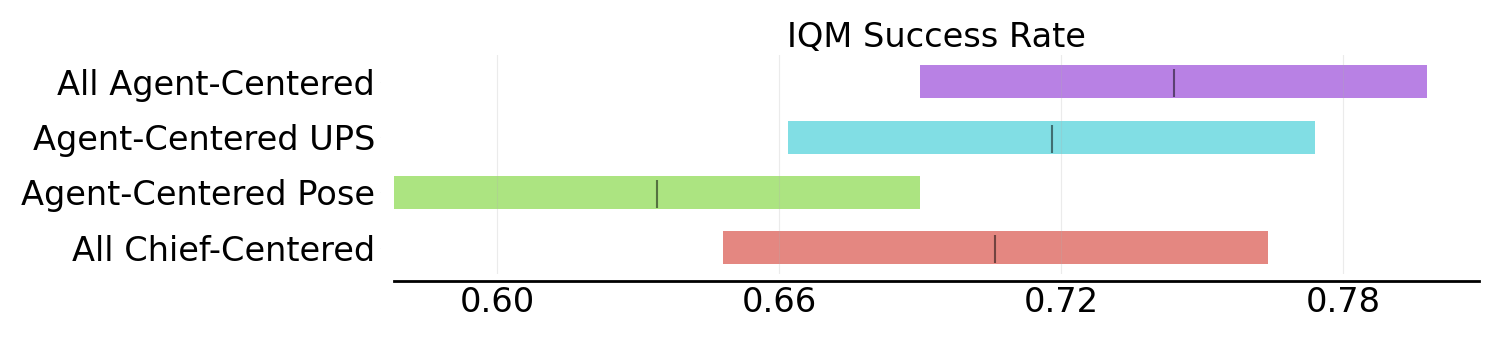}} \quad
    \subfigure[]{\includegraphics[width=.8\linewidth]{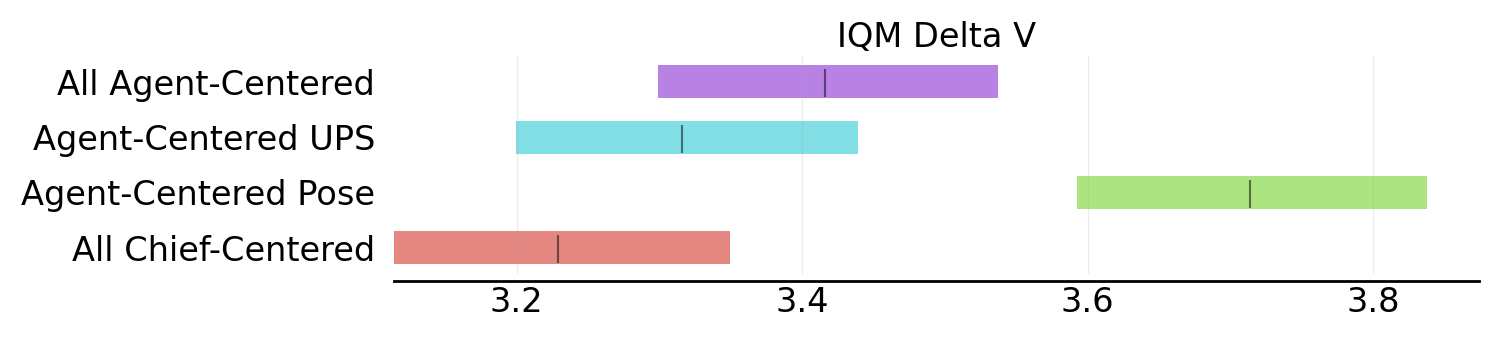}}
    \caption{Final policy performance plots showing the expected level of performance of the final trained models. The line represents the interquartile mean and the shaded region is the 95\% confidence interval.}
    \label{fig:ref_final}
\end{figure}

\begin{table}[]
\centering
\caption{Table of InterQuartile Means and 95\% CI for Reference Frame Study}
\label{tab:ref_frames}
\resizebox{1.0\linewidth}{!}{
\begin{tabular}{llllll}
\toprule
Metric Labels &       Total Reward &      Inspected Points &           Episode Length &       Success Rate &            Delta V  \\
\midrule
All Chief-Centered  &  9.32 [9.27, 9.36] &  98.26 [97.84, 98.55] &  342.37 [337.06, 347.92] &  0.71 [0.65, 0.76] &  3.23 [3.11, 3.35]  \\
Agent-Centered Pose &  9.28 [9.24, 9.31] &   98.1 [97.75, 98.37] &   338.51 [333.9, 343.67] &  0.63 [0.58, 0.69] &  3.71 [3.59, 3.84]  \\
Agent-Centered UPS  &  9.38 [9.35, 9.41] &   98.5 [98.22, 98.68] &   339.1 [334.36, 344.27] &  0.72 [0.66, 0.77] &   3.32 [3.2, 3.44]  \\
All Agent-Centered  &   9.37 [9.34, 9.4] &   98.6 [98.39, 98.74] &  357.51 [351.13, 364.36] &   0.74 [0.69, 0.8] &   3.42 [3.3, 3.54]  \\
\bottomrule
\end{tabular}
}
\end{table}

The results from training RL agents with the four reference frame configurations are shown in \figref{fig:ref_sample_complexity}, \figref{fig:ref_final}, and \tabref{tab:ref_frames}. The final performance metrics in \figref{fig:ref_final} and written out in \tabref{tab:ref_frames} show the differences between reference frames were not large enough to cause significant changes in behavior. 

However, the sample complexity results in \figref{fig:ref_sample_complexity} show the main advantage of using the agent-centered configurations. Within the first 2 million timesteps, the agent-centered configurations, particularly those with the agent-centered UPS, see a slight increase in reward because the policies are more fuel efficient. The slight advantage to the agent-centered configurations is overcome by 2 million timesteps. At that point, the RL agents have likely learned a similar transformation, or some other combination of weights in the NNC compensates for it. The NNC is able to compensate and/or learn the transformation quickly because the transformations are linear, making them easy to learn.

While the difference is small, repeating this experiment with a more complicated six-degree-of-freedom inspection task, which requires controlling rotation, could accentuate the trend. In this case, the agent-centered reference frame rotates as the agent changes its orientation with respect to the chief-centered reference frame. This creates much more complex transformations between the two reference frames, which would take longer for the NNC to learn and could make the difference between a successful and unsuccessful policy.
Initial experimentation in \cite{dunlap2024run} shows the agent-centered reference frame was successful for the task, but a full ablation study was not completed at the time.



\section{Conclusion}

The work in this paper reports analyses from training 110 unique agents to investigate how observation space design choices impact the learning process for space control systems.
In conclusion, the results show that (1) RL agents are capable of learning to successfully inspect the chief spacecraft in the illuminated scenario without the aid of any additional sensors, but the sensors do help the learning process and produce more optimal and consistent behavior. The results also revealed (2) that the \textit{count sensor}, which was designed to give the agent knowledge of how many points had been inspected so far, hinders the learning process and produces lower-performing policies. Finally, the results also showed that (3) which reference frame is used had minimal impact on the learning process for the translational inspection task.

Future work should consider a more complex, six-degree-of-freedom environment. A rotating agent might be more sensitive to changes in the reference frame when directional motion is dependent on orientation. Additionally, future work should consider multiagent environments, where information about other agents are added to the observation space.

\section*{Acknowledgments}
This research was sponsored by the Air Force Research Laboratory under the \textit{Safe Trusted Autonomy for Responsible Spacecraft} (STARS) Seedlings for Disruptive Capabilities Program.

The views expressed are those of the authors and do not reflect the official guidance or position of the United States Government, the Department of Defense, or of the United States Air Force. 
This work has been approved for public release: distribution unlimited. Case Number AFRL-2024-7047.

\bibliographystyle{AAS_publication}
\bibliography{references}









\end{document}